# Novel Convolution Kernels for Computer Vision and Shape Analysis based on Electromagnetism

Dominique Beaini[a*], Sofiane Achiche[a], Yann-Seing Law-Kam Cio[a], Maxime Raison[a]

[a]Polytechnique Montreal, 2900 Edouard Montpetit Blvd, Montreal, H3T 1J4, Canada

**Abstract**

Computer vision is a growing field with a lot of new applications in automation and robotics, since it allows the analysis of images and shapes for the generation of numerical or analytical information. One of the most used method of information extraction is image filtering through convolution kernels, with each kernel specialized for specific applications. The objective of this paper is to present a novel convolution kernels, based on principles of electromagnetic potentials and fields, for a general use in computer vision and to demonstrate its usage for shape and stroke analysis. Such filtering possesses unique geometrical properties that can be interpreted using well understood physics theorems. Therefore, this paper focuses on the development of the electromagnetic kernels and on their application on images for shape and stroke analysis. It also presents several interesting features of electromagnetic kernels, such as resolution, size and orientation independence, robustness to noise and deformation, long distance stroke interaction and ability to work with 3D images.

## 1. Introduction

Computer vision (CV) is a challenging and interdisciplinary field, with infinite possibilities of images and videos to be processed. Hence, it is not trivial to find the appropriate methodology to extract the desired data from the image. One of the favored approaches for image analysis is convolution kernels, which can be used for blurring, edge detection, defect detection, machine learning, etc. Therefore, creating new convolution kernels could unlock new possibilities of image processing or improve the existing methods.

Choosing the right convolution kernels for a task requires a mathematical understanding of each kernel, and may prove to be a tedious task. To avoid such problems, many methods rely on numerical optimization of the kernels, such as genetic algorithms [1, 2] or the highly-praised convolutional neural networks (CNN) [3–5]. This usually removes the need to create new types of kernels, since the optimized kernel will often be more efficient than a manually chosen one. However, electromagnetic (EM) kernels possess interesting features that can be used for computer vision, but that cannot be retrieved using standard kernel optimization. The current paper will explain how to build EM kernels, how to apply them and how to extract the useful information of shapes and strokes.

---

[*] Corresponding author. Tel.: +1-514-340-4711 # 3345
  *E-mail address:* dbeaini.phd@outlook.com

Click here to enter text.

**Nomenclature**

| | |
|---|---|
| $e, m$ | Electric ($e$) or Magnetic ($m$) |
| $dip$ | Dipole |
| $onC$ | Only values on the Contour |
| $I$ | Image matrix, with values between -1 and +1 |
| $E_{e,m}$ | Virtual Vector field $[V^0 \text{ pix}^{-1}]$ |
| $V_{e,m}$ | Virtual Potential $[V^0]$ |
| $P_{e,m,dip}$ | Virtual Potential kernel of a monopole or dipole $[V^0]$ |
| $q_{e,m}$ | Virtual Charge |
| $r$ | Virtual distance from an electric charge [pix] |
| $n$ | Number of spatial dimensions for the virtual potential |
| $\Delta^{x,y}$ | Numerical derivative kernel in $\hat{x}$ or $\hat{y}$ direction $[\text{pix}^{-1}]$ |
| $\mathcal{E}_{e,m}$ | Vector field $[V\,m^{-1}]_e$, $[V\,s\,m^{-2}]_m$ |
| $\mathcal{V}_{e,m}$ | Potential $[V]_e$, $[V\,s\,m^{-1}]_m$ |
| $\nabla$ | Gradient operator |
| $*$ | Convolution operator |
| $\circ$ | Hadamard product (Element-wise multiplication) |
| $\Re(), \Im()$ | Real part, Imaginary part |

*1.1. Related Work*

Convolution kernels are one of the favored method for extracting information from an image, and they have been the subject of numerous research works in noise removal [6], defect detection [7, 8], image segmentation [9, 10], edge detection [11], machine learning [3, 12, 13], etc. The convolutions allow to quickly scan the whole image to apply local mathematics on nearby pixels, either to extract features or to modify the image [4]. A common characteristic of the convolution kernels of all these methods is that the kernels are usually small in the bi-dimensional plane, with a maximum size of 7×7 [11], 9×9 [7], 11×11 [3], although some physical phenomena require bigger kernels, twice the size of the image [9, 10].

The current paper is not the first one to base its convolutions on natural phenomenon for the purpose of CV. For the biology inspired techniques, the most prominent one is neural networks (NN) based, more specifically the convolutional neural networks (CNN) [3, 4, 14]. This technique mimics how the human brain works by combining multiple sequences of neurons for the sake of classification or learning, and they are currently among the very best techniques for image classification [3, 13–15].

Some examples of physics inspired techniques for CV include watershed droplets for edge detection [16] and force vector fields used for active contours [9, 10], but they are not related to electromagnetism which is presented in this paper. The main methods related to the work presented in this paper are the quadrupole convolution used to define the orientation of contours [17] and the edge detection using gravitational fields (which is mathematically similar to electric fields) [11]. Although these 2 works initially proved to be efficient new ways of analysing images, they did not realize the full possibilities of using electromagnetism (EM). Therefore, these methods are now outranked by more recent techniques. The current paper aims to go a lot further in exploring the CV possibilities of electromagnetic potentials and fields (EMPF), such as oriented dipoles, interactive segments and combined potential-field analysis. It aims at developing a methodology of image analysis that directly uses the laws of electromagnetism as described by J.C. Maxwell [18–20].

Furthermore, the current paper deals extensively with the problems of shape and stroke analysis, which are important in the field of CV [21]. A stroke is defined as any curve in an image with a single pixel width. Some basic techniques are focused on giving general information about the shapes, such as the perimeter, area, centroid and mean size. These techniques are usually robust to deformations, since they use each point of a shape for computation, but they do not provide enough information for an advanced analysis [21], since they reduce a 2D shape into a 0D value. Other techniques transform a shape into its contour (either polygonal or smooth) [9, 21–24] or skeleton [21, 22, 25], hence transforming a 2D shape into a simpler 1D stroke. Those strokes are then analysed using their local curvatures, intersections or Fourier descriptors [9, 21, 22, 24–27]. Those techniques provide more information then the 0D values since they reduce the dimensionality by a lower factor, but they still greatly reduce the information initially available in the shape.

*1.2. Proposed Approach: CAMERA-I*

The objective of the proposed approach is divided into 2 parts. The first objective is to develop the mathematics and the methodology required to build and use the EM convolution kernels. The second objective is to explain how the kernels can be used in CV, and to demonstrate their advantages for shape and stroke analysis.

To answer the first objective, the EM laws are simplified to their static and multi-dimensional interpretation, and transformed into convolution kernels. Therefore, the EM potentials and fields of an image can be computed with simple convolutions. By analyzing their values and by determining the attraction or repulsion, it is possible to find several local or global characteristics of the images or shapes. The novel technique proposed in this paper is called "Convolution Approach of Magnetic and Electric Repulsion to Analyze an Image" (CAMERA-I).

For the second objective, this work aims at demonstrating that there are several advantages of the CAMERA-I approach, when compared to standard CV methods. One major advantage is the resolution and size independence of the kernels, which is something that is not possible with most other filtering methods [5, 28]. Another important characteristic is the high robustness to noise and deformation of the images. In fact, the proposed approach will prove to be way more robust to local changes, since the EMPF will consider the contribution of each pixel in the image, shape or stroke that is analyzed. Contrarily to standard methods, the proposed EMPF does not reduce the dimensionality of the shapes it analyzes, and even allows to analyze 3D objects. Finally, the EMPF approach will show how to extract long distance information about the strokes and shapes present in an image, allowing to consider the interaction between them and to build a 2D information image from a 1D stroke.

**2. Theory of Electromagnetism for Computer Vision**

In order for the paper to be self-explanatory, we use this section to remind the reader of the key concepts of electromagnetism (EM) that are useful to understand the paper. It will only deal with the classical theory of EM by J. C. Maxwell [18] and does not deal with relativity or quantum physics. We will then derive some simplified equations of EM, because computer vision deals with a virtual world, and is not subject to conform to the constants of the universe. The laws are modified to only consider the static terms, to remove the constants and to be used in a non-3D world, thus allowing to fully harness the geometrical properties of Maxwell's equations (MEq) without any physical limitation.

*2.1. Intuitive exercise*

For a better understanding of how electromagnetism can help determine features of an object, one can consider the physics of a lightning rod. It is well known that lightning will tend to fall on a sharp (highly convex) and tall (far from the center of mass) area, because of the high charge concentration near a sharp point [19]. Consequently, we can determine which regions are concave or convex, and which regions are near or far from the center of mass (CM). It is to note that the current paper goes far beyond this simple intuition and uses both mathematical and/or graphical evidence when required.

## 2.2. Electric and Magnetic Monopoles and Dipoles

Electromagnetic charges and dipoles are the key elements behind the CAMERA-I algorithm, because parts of the images are treated as EM particles. Such particles generate scalar potentials $\mathcal{V}$ and vector fields $\mathcal{E}$ that will vary according to the distance and the position [18–20]. For the sake of concision, most of the theory is given at "Appendix I Monopoles and Dipoles". We ignore the fact that magnetic monopoles do not exist, and we suppose that they behave identically to electric monopoles.

### 2.2.1. Electric or magnetic monopoles

The effect of electric monopoles is shown at Fig. 1 on a normalized color-scale, where $\mathcal{V}$ is represented by the color-scale and $\mathcal{E}$ is represented by the vectors. We see that the positive monopole produces a positive potential and outgoing field, while the negative particle produces a negative potential and an ingoing field.

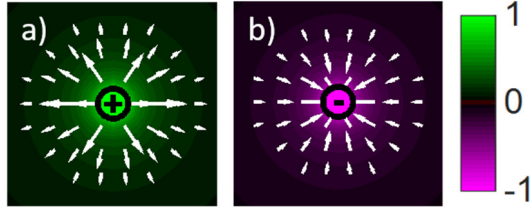

**Fig. 1.** Static electric potential and field of a: (a) positive monopole. (b) negative monopole

When multiple monopoles are put together, then the total potential is given by the scalar sum and the total field is given by the vector sum, as given by equation (1) [18–20].

$$\mathcal{V}_e^{tot} = \sum_i^n \mathcal{V}_e^i, \qquad \mathcal{E}_e^{tot} = \sum_i^n \mathcal{E}_e^i \qquad (1)$$

### 2.2.2. Electric or magnetic dipoles

When 2 monopoles of opposite signs are placed near each other, it produces a dipole with a potential and field that respect equation (1). Those dipoles can be stacked in several different ways, as observed on Fig. 2. It is shown that the serial assembling of dipoles does not make it stronger, it only creates a bigger gap between the negative and positive pole. However, placing the poles on 2 parallel lines will create a big dipole with a higher potential and field.

Another important characteristic is that when the distance between the poles is small, a dipole in any orientation $\theta$ is approximated by equation (2) [19, 20], where the superscripts x, y denote the horizontal and vertical orientation of the dipoles. A visual of this superposition is given at Fig. 2, where it is shown that a horizontal dipole with a vertical dipole is equivalent to 2 dipoles placed at 45°.

$$\mathcal{V}_{dip}^{\theta} \approx \mathcal{V}_{dip}^{x} \cos(\theta) + \mathcal{V}_{dip}^{y} \sin(\theta) \qquad (2)$$

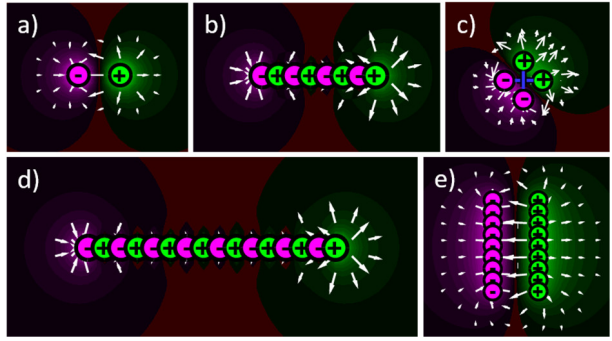

**Fig. 2.** Electric Potential and field for static monopoles placed as (a) a simple dipole. (b) a small chain of simple dipoles. (c) a horizontal and a vertical dipole, equivalent as 2 dipoles at 45°. (d) a long chain of simple dipoles. (e) simple dipoles in parallel.

*2.3. Potential and Fields Equations Adapted for Computer Vision*

In order to use the laws of EM, they must first be adapted for computer vision by removing some of the physical constraints and by ignoring the universal constants. In the "Appendix B. Mathematical Laws of EM", Maxwell's equations are presented and simplified using the assumption that all charges are static and that magnetic monopoles can exist. This allows to generalize the potential and field equations in a universe with $n$ spatial dimensions, where $n$ is a real number greater than 1. The modified field is presented at equation (3).

$$\boldsymbol{E}_{e,m} = q_{e,m} \frac{\hat{\boldsymbol{r}}}{|\boldsymbol{r}|^{n-1}}, \qquad n \in \mathcal{R}^+ \,\&\, n \geq 1 \tag{3}$$

By using the electromagnetic laws presented in the appendix, we can write the relation between the potential $V$ and its gradient $E$ at equation (4).

$$\boldsymbol{E}_{e,m} = -\nabla V_{e,m}$$
$$V_{e,m} = -\int_C \boldsymbol{E}_{e,m} \cdot \mathrm{d}\boldsymbol{l} \tag{4}$$

It is then possible to determine the potential by calculating the line integral of equation (3). This leads to equation (5), where we purposely omit all the integral constants and the product constant terms that depends of $n$.

$$V_{e,m} \propto q_{e,m} \cdot \begin{cases} |\boldsymbol{r}|^{2-n}, & n \geq 1,\ n \neq 2 \\ \ln|\boldsymbol{r}|, & n = 2 \end{cases} \tag{5}$$

For $n = 3$, $V_{e,m} \propto |\boldsymbol{r}|^{-1}$, which is identical to the real electric potential in 3D [18–20]. Because the field is the gradient of the potential, then the vector field will always be perpendicular to the equipotential lines, and its value will be greater when the equipotential lines are closer to each other [19].

For the current paper, the term "**electric**" is used when using monopoles and "**magnetic**" or "**magnetize**" when using dipoles, because it is more intuitive.

*2.4. Geometrical Interpretation of Potentials and Fields*

The mathematical formalism of EM and MEq have been presented and simplified, but with no purpose for shape analysis, which will be the main focus of this section. An in-depth analysis of circles and corners are presented at "Appendix C. Geometrical Interpretation of Maxwell's Equations" and can help understand the following interpretation.

If a given shape is filled of positive electric monopoles, then the field will tend to cancel itself near the center of mass (CM) or in concave regions. However, the potential is scalar, which means that it will be higher near the CM or in concave regions. This difference in the behavior of the potential and the field is observed at Fig. 3.

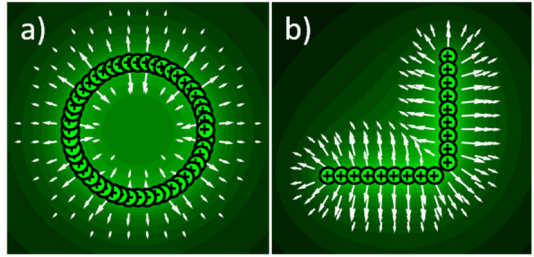

**Fig. 3.** Potential and field with $n = 3$ for positive monopoles placed on (a) A circle. (b) A corner.

Using this difference, we can determine the features of the shape in a given region depending only on the values of $V_e$ or $|E_e|$. The characteristics of the potential and the field in different regions of the shape are summarized at Table 1. Of course, a combination of those factors is possible, like a concave region near the center of mass (CM), which yields to a really high potential and a slightly low field.

Table 1 : Potential and Field Characteristics at Different Regions of a Shape Filled of Monopoles, for $n > 2$

| Region of interest | Visual | $V_e$ | $|E_e|$ | Legend | |
|---|---|---|---|---|---|
| Concave | | ↑ | ↓ | ↓↓ : | really low |
| Convex | | ~↓ | ~↓ | ↓ : | low |
| Flat | | ~ | ↑ | ~↓ : | slightly low |
| | | | | ~ : | average |
| Near CM | | ↑ | ~ | ~↑ : | slightly high |
| Far from CM | | ↓ | ↓ | ↑ : | high |
| Inside | | ↑↑ | ↓↓ | ↑↑ : | really high |

## 2.5. Convolutions, Potentials and Fields

The equations (4) and (5) are the main equations used in this paper. The potential is first calculated using equation (5) because it represents a scalar, which means it is easy to sum the contribution of every monopole by using 2D convolutions. Then, the vector field is calculated from the gradient of the potential. Convolutions are used because they are fast to compute due to the optimized code in some specialized libraries such as *Matlab®* or *OpenCV®*.

### 2.5.1. Creating the monopole potential kernel

Knowing that the total image potential is calculated from a convolution, the first step is to manually create the potential of a single particle on a discrete grid or matrix. The matrix must be composed of an odd number of elements, which allows to have one pixel that represents the center of the matrix. If the size of the image is $N \times M$, it is preferable to have $P_e$ as a matrix of size $(2N + 1) \times (2M + 1)$. This avoids having discontinuities in the potential and its gradient. However, it means that the width and height of the matrix can be of a few hundred elements and take a longer time to compute without efficient libraries. The convolution kernel matrix for $P_e$ is calculated the same way as $V_e$ at equation (5), because it is the potential of a single charged particle, with the distance $r$ being the Euclidean distance between the middle of the matrix and the current matrix element. An example of a small $P_e$ matrix of size $7 \times 7$ is illustrated in Fig. 4, where it is noted that $P_e$ is forced to "1" at the center for continuity purpose.

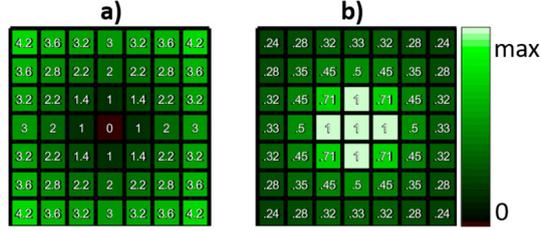

**Fig. 4.** Example of convolution kernel for a particle potential matrix $P_e$ of size $7 \times 7$; (a) Euclidian distance from center $r$. (b) Potential of a centered monopole $P_e = V_e$, $n = 3$.

### 2.5.2. Creating the dipole potential kernel

Convolutions with dipole potentials can be used also to create an anti-symmetric potential and find the specific position of a point. Therefore, it is required to create a potential convolution kernel for a dipole $P_{dip}$. We have to remember that a dipole is simply 2 opposite monopoles at a small distance from each other. This can be expressed as a mathematical convolution where $P_{dip}$ is given by equation (6), and is visually shown in Fig. 5. If divided by a factor 2, we can notice that this convolution is similar to a horizontal numerical derivative (shown later at equations (8) and (9)), meaning that the dipole potential is twice the derivative of the monopole potential [19].

$$P_{dip}^x = P_e * [-1 \quad 0 \quad 1], \qquad P_{dip}^y = -(P_{dip}^x)^T$$
$$\text{size}(P_{dip}) = \text{size}(P_e) \tag{6}$$

Using equation (2) along with equation (6), it is possible to determine equation (7), which gives the dipole kernel at any angle $\theta$.

$$P_{dip}^\theta \approx P_{dip}^x \cos(\theta) + i\, P_{dip}^y \sin(\theta) \tag{7}$$

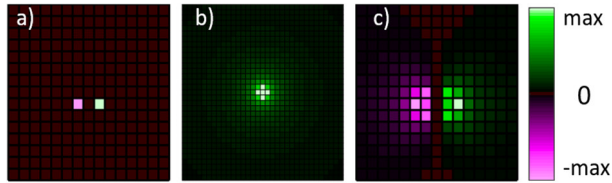

**Fig. 5.** Steps to calculate the normalized potential kernel for a dipole (a) Positive and negative monopoles at 1 pixel distance. (b) Potential kernel $P_e$. (c) Dipole potential kernel $P_{dip}^x$ resulting from the convolution of image "a" with kernel "b".

### 2.5.3. Creating the derivative kernels

Derivative kernels are important to calculate the field, because we know from equation (4) that the field $E_{e,m}$ is the gradient of the potentials $V_{e,m}$. To use the numerical central derivatives, we simply need to apply the convolution given at equation (8), with the central finite difference coefficients given at equation (9) for an order of accuracy (OA) of value 2 [29].

$$\frac{df}{dx} \approx f * \Delta^x, \qquad \frac{df}{dy} \approx f * \Delta^y \tag{8}$$

$$\Delta^x = (\Delta^y)^T = \frac{1}{2}[-1 \quad 0 \quad 1], \qquad OA = 2 \tag{9}$$

*2.5.4. Calculating the potential and the field of an image*

A crucial step for the CAMERA-I technique is to transform an image into charged particles, which will allow calculating the electric potential and field. The first step is to determine the position and intensity of the charge. Each pixel with value $+1$ is a positive monopole, each pixel with value $-1$ is a negative monopole, and each pixel with value 0 is empty space. Therefore, the pixels of the image represent the density of charge and have values in the interval $[-1, ..., 1]$, where non-integers are less intense charges.

Next, the $P_e$ matrix is constructed as seen on Fig. 4, and applied on the image with the convolution shown at equation (10). Then, the horizontal and vertical derivatives are calculated using equation (8) and give the results for $E^x$ and $E^y$. Finally, the norm and the direction of the field are calculated using equation (11). It is possible to visualize these steps at Fig. 6, where a quadrupole is represented.

$$V_e = I * P_e, \quad \text{size}(V_e) = \text{size}(I) \tag{10}$$

$$E^{x,y} = V_e * \Delta^{x,y}$$

$$|\mathbf{E}| = \sqrt{(E^x)^2 + (E^y)^2} \tag{11}$$

$$\theta_E = \text{atan2}(E^y, E^x)$$

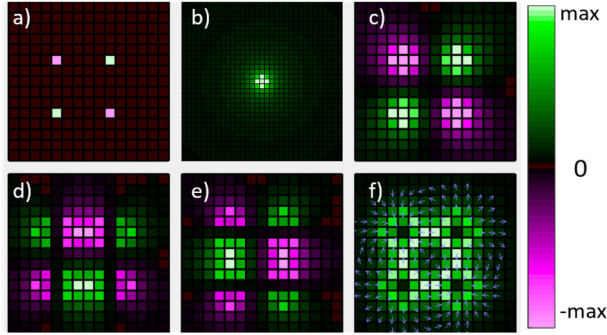

**Fig. 6.** Calculation of the potential and field of an image (a) Monopoles in the image. (b) Potential kernel $P_e$. (c) Total potential $V_e$. (d) Horizontal field $E_e^x$. (e) Vertical field $E_e^y$. (f) Field norm $|\mathbf{E}_e|$ and direction

The same process that is used to transform each pixel into a monopole can be used to transform them into a magnetic dipole, by using the result presented at Fig. 5 as the kernel. The steps and results are shown at Fig. 7, when each pixel is transformed into a horizontal magnetic dipole with $\theta = 0$. The formula to calculate the magnetic potential using a convolution is given at equation (13), with the density correction factor $F$ shown at equation (12). This density factor $F$ allows to consider the fact that pixels placed in diagonal have a lower number of pixels per unit length then those placed horizontally. The angle $\theta$ depends on the image, as it is often chosen to be either parallel or perpendicular to the direction of the element to magnetize. Also, the matrix size of $V_m$ is the same as the matrix size of $I$. The real part is chosen in equation (13) to represent dipoles perpendicular to $\theta$, while the imaginary part represents dipoles parallel to it.

$$F = \max(|\cos(\theta)|, |\sin(\theta)|)^{-1} \Rightarrow 1 \leq F \leq \sqrt{2} \tag{12}$$

$$V_m = \Re\left(\left(I \circ F \circ e^{i\theta}\right) * P_{\text{dip}}^\theta\right) \tag{13}$$

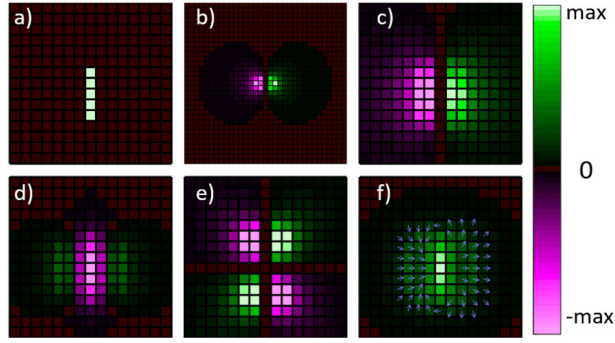

**Fig. 7.** Steps to calculate the magnetic PF of an image (a) Dipoles in the image. (b) Horizontal dipole potential kernel $P_m^x$. (c) Total potential $V_m$. (d) Horizontal field $E_m^x$. (e) Vertical field $E_m^y$. (f) Field norm $|\boldsymbol{E}_m|$ and direction.

It is to note that the image (e) of Fig. 7 is really similar to the image of quadrupole potential presented at Fig. 6. This is because it represents 2 consecutive perpendicular derivatives of the potential of monopoles $\partial/\partial x (\partial/\partial y V_e)$, which is mathematically equivalent to a quadrupole potential.

## 3. Application of EM Convolutions

The previous section explained how to correctly build the convolution kernels, although real kernels are a lot bigger than the schematic demonstrations. In this section, the focus will shift on how to use those kernels for shape and stroke analysis, and what are the advantages of using EM convolution kernels.

### 3.1. Detecting Shape Characteristics

This first sub-section will focus on the use of EM convolution kernels for the detection of multiple shape characteristics, such as the convex or concave regions, and the relative distance to the centroid.

### 3.1.1. Finding the regions of interest

It was discussed in the section "2.4 Geometrical Interpretation of Potentials and Fields" that the electric potential and field can be used for shape analysis, with a summary of the characteristics presented at Table 1. To demonstrate those characteristics, a special shape is created with all the mentioned regions of interest (RoI), with the computed potential and field shown at Fig. 8. The index "onC" means that the values were set to 0 everywhere but on the contours. It is to note that the PF are computed using the whole surface of the shapes, and that the values are set to 0 after the computation of $V_e$ and $|\boldsymbol{E}_e|$. The contour of a full shape can be easily determined with morphological operations. The values of the potential $V_e^{\text{onC}}$ and field $|\boldsymbol{E}_e^{\text{onC}}|$ on the contours are squared to show a better contrast between the low values and the high values. They are also thickened using image dilation, for the purpose of showing better images.

The value of the dimension is set to $n = 3$ for these examples, as it is found experimentally to be ideal for such analysis. By choosing a value of $2 < n < 3$, a similar interpretation can be done, but it will increase the contribution of the pixels very far from each other, and significantly reduce the contribution of nearby pixels. By choosing a value of $n > 3$, it will reduce the contribution of pixels that are far from each other, and increase the contribution of nearby pixels. Hence, the value of $n = 3$ was found to be a good equilibrium of the contribution of nearby and far pixels, although each specific application could optimize its value. For a more advanced analysis, it is possible to use various different dimension values, such as $n = \{2.3,\ 3,\ 4\}$, and to combine the information given by each value of $n$.

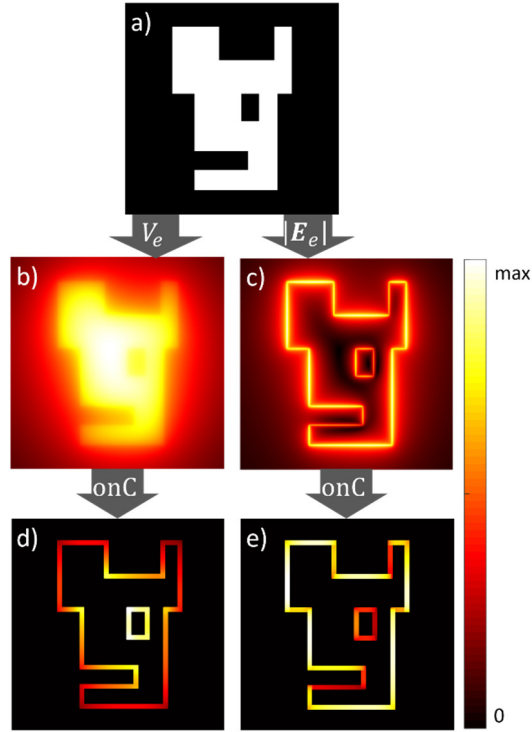

**Fig. 8.** (a) Special shape with the white region being a uniform density of charge, used to compute the following PF with $n = 3$. (b) The potential $V_e$. (c) The field $|E_e|$. (d) The potential squared only on the contour $\left(V_e^{\text{onC}}\right)^2$. (e) The field squared only on the contour $\left|E_e^{\text{onC}}\right|^2$.

Using the values of $V_e^{\text{onC}}$ and $\left|E_e^{\text{onC}}\right|$ depicted at Fig. 8, it is possible to find the regions of interests, as seen at Fig. 9. The percentile thresholds that are used are shown in Table 2. Since the shape that is used is complex, the regions are not perfectly discernable, as usually expected. For example, a concave region (which expects a high value of $V_e$) can also be far from the CM (which expects a low value for $V_e$), hence, the thresholds are contradictory. However, this can be used as an advantage, since it allows to use general information about an image, and make it more robust to noise. In fact, regular convolution kernels are small, which makes them vulnerable to small variations in the shapes contours, but it is not the case for EM kernels. EM kernels are also invariant in rotation and robust to deformations.

It is to note that using simple thresholds might lead to problems regarding the continuity of a region, which could be fragmented in a few small parts. To avoid this problem, all regions are grown by a security factor, which is chosen as 5% of the biggest dimension of the shape (this factor can be changed depending on the needs). Using such a percentage allows the growing to be robust, no matter the resolution or the size of the shape. The algorithm for such a region growing is explained in the Algorithm 2, at the Appendix "E.2. Grow and unite contour regions".

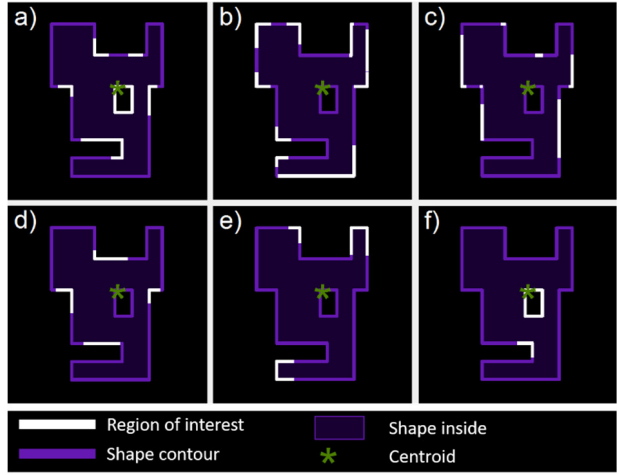

**Fig. 9.** RoI found on a complex shape using a contour analysis by potential and field thresholds. (a) Concave regions. (b) Convex regions. (c) Flat regions. (d) Regions near the CM. (e) Regions far from the CM. (f) Regions inside the shape.

Table 2. Percentile Thresholds Used for the Discovery of the Regions of Interest

| **Region of interest** | **Thresholds percentile for $V_e$** | | **Thresholds percentile for $|E_e|$** | |
|---|---|---|---|---|
| | Min (%) | Max (%) | Min (%) | Max (%) |
| **Concave** | 70 | 100 | 0 | 50 |
| **Convex** | 15 | 40 | 15 | 40 |
| **Flat** | 40 | 60 | 80 | 95 |
| **Near CM** | 80 | 95 | 40 | 60 |
| **Far from CM** | 0 | 25 | 0 | 25 |
| **Inside** | 90 | 100 | 0 | 10 |

*3.1.2. Robustness to deformation*

To demonstrate the robustness of the technique, the shape of Fig. 9 is modified using a combination of the following filtering, both on small and big scale: twirl, twist and wave. The shape resulting from the filtering is presented at Fig. 10, with the RoI computed using once again the thresholds of Table 2. As it can be observed, the discovered regions for both Fig. 9 and Fig. 10 are almost identical, with only minor differences. From all the RoI, the only differences comprise of one convex region, one flat region and one region far from the CM. All other regions are present on both figures at the same place. Those differences are minor and are expected since the shape has been greatly modified by the multiple filtering.

Hence, we show that the proposed technique is highly robust against shape and contour deformation for detecting RoI. This is mainly due to the electric field that considers every pixel inside a shape, not only those in a small region of the contour. In fact, although the contour of the shapes on Fig. 9 and Fig. 10 are greatly different, the total pixels inside the shape area had a lot less variation, which means that the values of the EMPF are almost identical. Other convolution kernels are small, meaning that they focus only on local information. Hence, kernels that detect concave regions will detect any bump in the contour that is locally concave, making it really vulnerable to deformation.

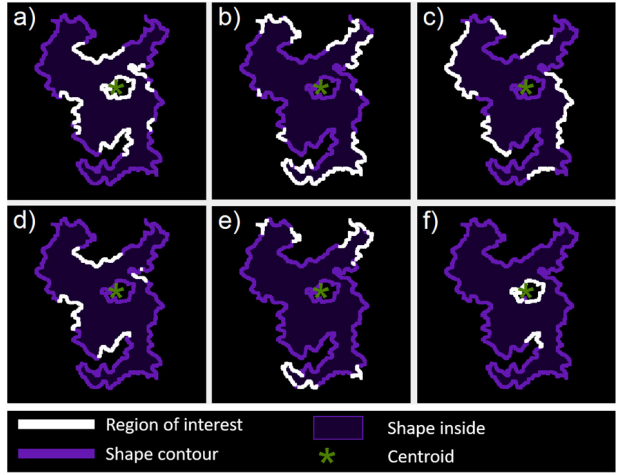

**Fig. 10.** RoI found on a complex shape (filtered with twirl, twist and wave distortion) using a contour analysis by potential and field thresholds. (a) Concave regions. (b) Convex regions. (c) Flat regions. (d) Regions near the CM. (e) Regions far from the CM. (f) Regions inside the shape.

*3.1.3. Analysis of 3D shapes*

In spite of being robust to deformation, an important characteristic of EMPF is that they can be used in 3D without any added complexity. Of course, computation time will be longer in 3D, but it is partly compensated by resolution which is usually lower than 2D images.

However, since there is more information in 3D than 2D, it is easier to analyse a 3D objects using different values of $n$. The same rules of Table 1 still apply, but a value of $n < 4$ will be more sensible to the CM, while a value of $n > 4$ will be more sensible to the local convexity. An example of result for a 3D mug is presented at Fig. 11, with $n = \{3, 4\}$. It can be observed that $|E_{onC}|_{n=3}$ is better at determining the inside of a cup with opposing faces, while $|E_{onC}|_{n=4}$ is better at finding the bottom of the cup, where the concavity is the highest. Furthermore, $(V_{onC})_{n=4}$ is better than $(V_{onC})_{n=3}$ at finding the local convexities at the border of the cup.

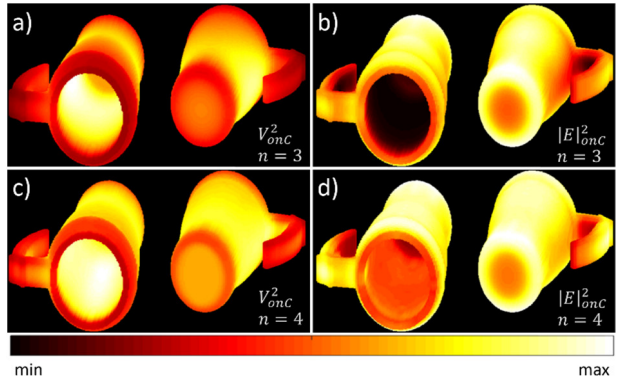

**Fig. 11.** EM potential $V$ and field $E$ generated by a 3D mug, with different values of $n$. (a) $V_{onC}^2$ with $n = 3$. (b) $|E|_{onC}^2$ with $n = 3$. (c) $V_{onC}^2$ with $n = 4$. (d) $|E|_{onC}^2$ with $n = 4$.

## 3.2. Magnetic Repulsion for Stroke Interaction

As demonstrated in the last section, the electric potential and field allow to analyze a shape and its contour. In this section, a new tool will be developed to show how magnetism can be used to analyze thin strokes and their interactions. A thin stroke is defined as any curve or line that has only one-pixel width. Hence, each pixel of the stroke has a maximum of 2 neighbors, except at the intersection of multiple strokes.

*3.2.1. Choosing the Magnetic Dimension*

One major difference between analyzing full shapes and strokes is the impact of image resolution. For a full shape, if the resolution is lowered, the total relative area between the shape and the image remains the same.

For a thin stroke, if the resolution is lowered, then each pixel of the stroke is wider. Hence the stroke has a bigger relative area when the resolution is low. This causes problems when using EM convolutions, since the area represents the total charge. However, it is found that using $n = 2$ for the EMPF makes it invariant of the thickness of the stroke and the resolution of the picture. This can be observed at Fig. 12, where 2 strokes of different resolutions are magnetized perpendicular to the stroke with dimensions $n = 2$ and $n = 3$. For the value of $n = 2$, presented at the subfigures (c) and (f), we can observe that the equipotential lines are exactly the same. However, this is not the case for subfigures (b) and (e), where $n = 3$. The potentials of Fig. 12 are computed using equation (13), with $\theta$ being the orientation perpendicular to the line. To find the orientation $\theta$ for any kind of stroke, it is possible to use Algorithm 3 in the Appendix "E.3. Finding Stroke Orientation".

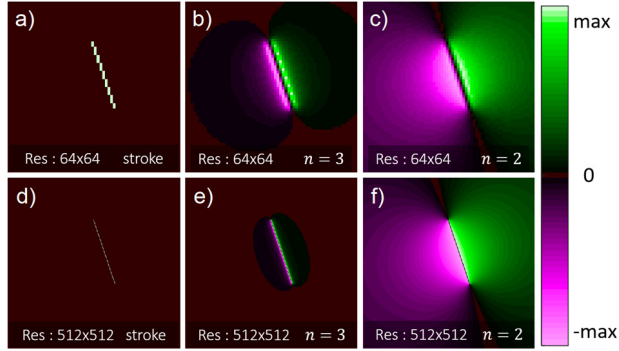

**Fig. 12.** Potential $V_m$ resulting of the convolution of a dipole perpendicular to the stroke lines. (a) Stroke with low resolution 64x64. (b) Dipole with $n = 3$ and low resolution. (c) Dipole with $n = 2$ and low resolution. (d) Stroke with high resolution 512x512. (e) Dipole with $n = 3$ and high resolution. (f) Dipole with $n = 2$ and high resolution.

Another important characteristic of the potential with $n = 2$ is that it is the only dimension which ensure a conservation of energy in the potential and field of the image, since the image is in 2D. In fact, the conservation of energy is the reason why a thin stroke requires $n = 2$ to be invariant of the image resolution. If we chose a value of $n > 2$, then some energy will be lost in the higher dimensions as we go further from the EM particles. Inversely, a value of $n < 2$ will create more energy as we go further from the EM particles. Using the same principles, it is possible to deduct that a thin plane in a 3D image requires $n = 3$ to be invariant of the resolution.

The conservation of energy means that Gauss's Theorem can be applied on the field produced by a stroke. By using Gauss's Theorem, we can know that any closed stroke, which is magnetized perpendicular to its direction, will produce a null field both inside and outside the stroke. The fact that the field is null means that the potential is constant, both inside and outside the stroke, but with different values. This can be observed at Fig. 13, where the closed stroke is chosen to be a circle. The more the circle is near closing, the more the potential is uniform. However, this is only true for $n = 2$. The value of $V_m$ is given by equation (13), with $\theta$ computed using Algorithm 3.

In summary, the dimension value for the stroke analysis must be $n = 2$, for both purposes of resolution invariance and conservation of energy.

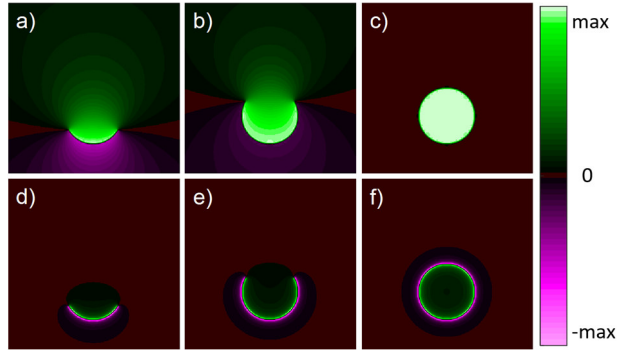

**Fig. 13.** Potential $V_m$ of a circular stroke magnetized perpendicular to their orientations. (a) Circle arc of 90°, with $n = 2$. (b) Circle arc of 270°, with $n = 2$. (c) Circle arc of 360°, with $n = 2$. (d) Circle arc of 90°, with $n = 3$. (e) Circle arc of 270°, with $n = 3$. (f) Circle arc of 360°, with $n = 3$.

*3.2.2. Magnetic Interaction*

As seen previously on Fig. 13 with $n = 2$, a stroke that is almost closed will have a higher potential $|V_m|$ inside it, with a lower potential outside. This can also be applied to 2 strokes that interact with each other by magnetizing them perpendicular to the strokes with equation (13). It is possible to shift the value of $\theta$ by a factor of $\pi$ on each stroke to flip the positive and negative side. By choosing carefully which stroke is flipped, it is possible to maximize the magnetic repulsion in an image, as shown at Fig. 14.

When there is a magnetic attraction, which is when the positive (green) part of a stroke meets the negative (pink) part of another stroke, nothing interesting happens in terms of the potential. However, when there is a repulsion (positive meets positive, or negative meets negative), there is a high concentration of potential $|V_m|$ between the strokes, with an almost constant value (low field $|E_m|$). Henceforth, the magnetic interaction is interesting, as it offers an opportunity to analyze the whole 2D space using only thin 1D strokes in the initial image.

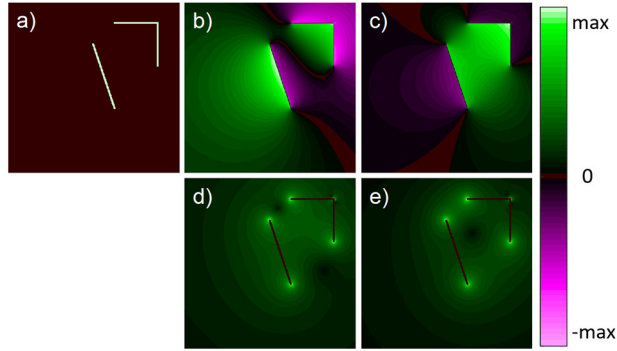

**Fig. 14.** PF computed from the initial stroke, with $n = 2$ and the dipole perpendicular to the strokes. (a) Initial stroke. (b) Potential of attraction $V_m$. (c) Potential of repulsion $V_m$. (d) Field of attraction $|E_m|^{0.5}$. (e) Field of repulsion $|E_m|^{0.5}$.

*3.2.3. Stroke Analysis*

Similarly to the problem of detecting shape characteristics using electric PF, presented in section "3.1 Detecting Shape Characteristics", it is possible to detect the characteristics of a stroke using magnetic PF. Furthermore, the stroke analysis will be robust to deformation, for the same reasons as the robustness of the shape analysis. To analyze a stroke, one must simply consider the potential $|V_m|$ produced by dipoles placed perpendicular to the stroke, using equation (13) and Algorithm 3. Then, as seen on Fig. 13, a concave region will produce a higher value of $|V_m|$, while a convex region will produce a lower value. This is also analogous to the magnetic repulsion interaction presented in the section "3.2.2 Magnetic Interaction". An example of this method and its robustness is presented in the Fig. 15, where it is observed that the values of $|V_m|^2$ are almost identical for the stroke, the deformed stroke and the heavily distorted stroke.

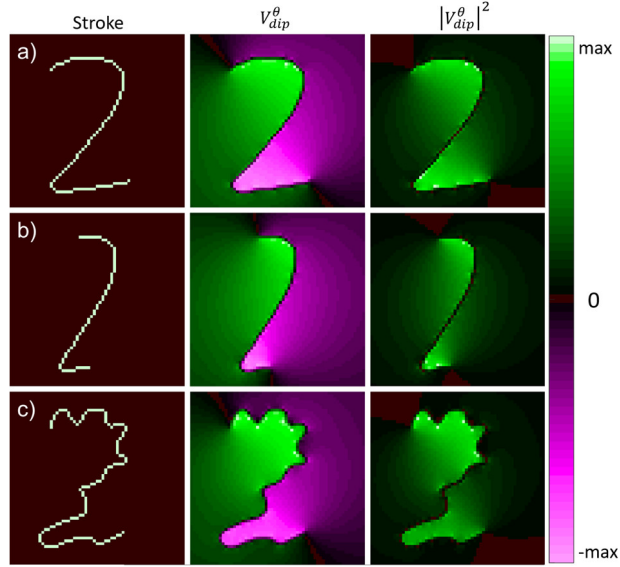

**Fig. 15.** Strokes for the number "2" at the left, with the potentials $V_m$ of dipoles perpendicular to the strokes, with $n = 2$. (a) Clean stroke. (b) Deformed stroke. (c) Heavily distorted stroke.

### 3.3. Summary of The Advantages of Electromagnetic Convolution Kernels

The focus of the current paper was mostly about the development and the appropriate usage of EM convolution kernels. Although many characteristics were presented, no concrete application is developed, making it harder to understand the real advantage of such an unusual approach. Hence, this section will focus on enumerating and explaining the great advantages and the uniqueness of EM for image analysis, when compared to other methods.

### 3.3.1. Resolution and Size Independence

The first clear advantage is the resolution and size independence of any EM convolution kernel. This characteristic is unique and is present both in shape and stroke analysis. This means that no matter the resolution of the image or the size of the shapes or stroke in the image, the ideal kernel size is $(2N + 1) \times (2M + 1)$, with $N$ and $M$ being the width and height of the image. If the resolution of the image is doubled, the kernel size is also doubled, but it won't change the results (although a super low resolution will be prone to numerical errors and might change the characteristics of the image). This is a characteristic that most kernels in the literature do not respect, since they are typically with a size between $3 \times 3$ and $31 \times 31$ [5, 28]. Hence, changing the resolution of the image or the elements inside the image requires to change the resolution of the kernels [5, 28]. This is problematic since the pixel width and height of a feature is unknown and is not necessarily dependent on the resolution of the image.

### 3.3.2. Orientation independence

In addition of being independent of size and resolution, the proposed EM kernels are also orientation independent, meaning that any rotation applied to the image will not alter the results. This is a feature that is usually available only with rotation symmetric kernels [5], such as Gaussian filters or the $P_e$ kernel presented at equation (5). However, the current method also presents how to use the asymmetric kernel $P_{dip}^\theta$ of equation (7) such that it is independent of the orientation. This is because the value of $\theta$ is dependent on the local orientation of a stroke, which changes along with the rotation of the image. Hence, what is presented is a unique asymmetric kernel that is independent in size, resolution and orientation. This is in contrast with other kernels, such as texture algorithms, which usually require 4 to 11 scales and 2 to 8 orientations [5, 30], for a total of 4 to 88 filters required for the same feature detection.

### 3.3.3. Robustness to Deformation and Noise

Another interesting feature is the robustness to heavy deformation, which was previously shown at Fig. 10 and Fig. 15. This is something that standard kernels cannot handle well due to their size. In fact, a small kernel will be way more affected by a local distortion, meaning that a standard kernel will find convex and concave regions almost everywhere in the presented distorted figures [30]. Furthermore, the standard techniques of contour approximation, such as the polygonal approximations and the Fourier descriptors, are too heavily affected by heavy variations on the contour [21, 26]. This is because those techniques rely only on the pixel of the contours, while the approach considers every pixel inside the shape. Those pixels inside the shape are far less affected by deformation than those on the contours. Similarly, adding noise to the pixels inside a shape will negligibly affect the total potential and field generated, since positive and negative noises will tend to even out.

### 3.3.4. Long Distance Interaction

An important feature of EM kernels is that they also allow to take into account the interaction between different strokes or shapes, as shown in Fig. 14. It potentially allows to group multiple strokes together, or the find the total PF generated by multiple shapes. This is a characteristic that is only possible using big kernels, since it is impossible for a small kernel to link 2 distant strokes. Other methods also propose a long distance interaction, such as the force vector fields for active contours [9, 10], as they use vector kernel to find the force interaction between each part of a contour.

### 3.3.5. Full Space Information

Another unique feature of the EM kernels is that they allow to analyze pixels that are not in the shape or stroke of interest. For example, the stroke analysis of Fig. 15 allow to tell if each pixel is positioned inside the concave regions of the number "2", or if it is in the convex region. This is impressive, since convolution kernels usually give only local information of an image. However, in that case, the EM kernels is able to generate 2D information from a 1D stroke.

### 3.3.6. Does not Require Shape Approximations

The EMPF approach has the unique capability of not reducing the dimensionality of the studied shapes. In fact, it was stated in the section "1.1 Related Work" that the other shape analysis techniques reduce the shape into a 0D value or a 1D contour/skeleton. These dimensionality reductions make the analysis simpler, but they tend to remove some critical information. Furthermore, techniques such as polygon approximation and Fourier descriptors require to approximate the shape of an object, which does not work well with complex shapes or shapes with holes [21, 23, 24].

Furthermore, EMPF even has the possibility of being used on 3D shapes, as seen at Fig. 11, since the laws of EM can still be applied using equation (5). Since 3D shapes are far more complex than 2D shapes, the advantage of using the proposed CAMERA-I approach is even greater, as it does not require any shape approximation.

### 3.3.7. Cannot be Learned

Since the dawn of CNNs, there is no point of creating a new convolution kernel if it can be learned by the network. The reason is that CNNs use dozens or hundreds of optimized kernels [3], meaning that any useful and "learnable" kernel will be obtained by the network optimization. However, the EM kernels presented in the current paper cannot be learned by such methods, and for several reasons. First, it was already mentioned that the kernels in a CNN are small [3, 4, 14], usually less than $11 \times 11$. Hence, it is impossible to learn a kernel that is twice the size of the image. Another important reason is the use of the magnetic potential $V_m$ that requires to convert the image into complex numbers using Euler's formula $\exp(i\theta)$, convoluted with a kernel of complex values, as seen in equation (13), with the angle $\theta$ being related to the direction of the stroke. This kind of specific feature is impossible to generate throughout the optimization of standard CNN, since they do not use complex numbers.

## 3.4. Comparison to the literature

The advantages presented at the previous section highlighted the interesting characteristics of the CAMERA-I approach for Computer Vision. A summary of these advantages is listed in Table 3, with a direct comparison to state-of-the-art methods of image analysis.

Table 3 : Qualitative Comparison Between Different Image Analysis Methods

| Characteristics | CAMERA-I | CNNs [3–5] | Fourier Descriptors [23, 24] |
| --- | --- | --- | --- |
| Resolution and size independence | ✓ | | ✓ |
| Orientation independence | ✓ | | ✓ |
| Robustness to deformation | ✓ | ~ | ✓ |
| Long distance interaction | ✓ | | |
| Full space information | ✓ | | |
| Avoids shape approximations | ✓ | ✓ | |
| Allows shape analysis | ✓ | ✓ | ✓ |
| Allows stroke analysis | ✓ | ✓ | |
| Can be adapted to any problem | | ✓ | |
| Can be easily used for machine learning | | ✓ | |
| Does not require heavy computing | ✓ | | ✓ |

From Table 3 one can see that the CAMERA-I approach is complementary to CNNs, but in direct competition with Fourier descriptors. One need to note that the examples of results using Fourier descriptors with 4 or 32 harmonics are illustrated in Fig. 16. First of all, we can clearly see that 4 harmonics is not enough to describe complex shapes. Using 32 harmonics, the results look better, but there is a lot of oscillations, which makes it difficult to accurately determine the convex and concave regions, since this technique relies on the local curvature. In addition, we can observe that Fourier descriptors cannot deal with holes in the shapes, and must consider the holes as separate shapes, contrarily to the CAMERA-I approach (as previously shown in Fig. 9 and Fig. 10). Finally, Fourier descriptors are only good to analyze full shapes, and cannot be used for stroke analysis or stroke interactions, which is another advantage of the proposed approach.

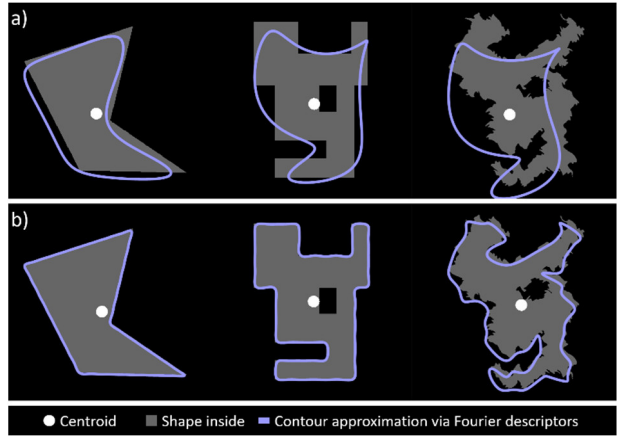

**Fig. 16.** Contour approximation via Fourier descriptors. (a) Fourier descriptor with 4 harmonics. (b) Fourier descriptors with 32 harmonics. [23, 24]

## 4. Conclusion

The objective of this paper was to develop different electromagnetic convolution kernels that can be used in computer vision applications, and to demonstrate its effectiveness for shape and stroke analysis. The paper showed how to express the images as electromagnetic particles with possible varying density, allowing to efficiently compute the potential $V_e$ and field $|E_e|$ associated to them, using convolutions. Using the computed values, it is possible to quickly determine some regions of interest, such as the convex or concave regions, and the proximity to the centroid. This method was demonstrated to be robust to noise and heavy deformation, and invariant to size, resolution and orientation.

Furthermore, a novel directional magnetic convolution is presented at equation (13), which allows to compute a $V_m$ and field $|E_m|$ that depend on the local density and orientation of thin strokes. This is a unique way of applying convolution kernels, which proved to be robust to heavy deformations, to allow high distance stroke interaction and to determine local or global stroke characteristics. Plus, it offers a unique way to analyze all the pixels in a 2D image depending on their relative position to the 1D stroke.

In summary, the electromagnetic kernels proved to be an efficient and robust way to analyze images, with unique characteristics that make it impossible to be the result of an optimized or learned kernel. In fact, the EM kernels proved to be independent of the image size, resolution or orientation, in addition of being really robust to deformation. A continuation of this work could focus on the development of specific applications based on those properties.

**Acknowledgment**

We would like to thank NSERC, through the discovery grant program, and FRQNT/INTER for their financial support as well as MEDITIS (Biomedical technologies training program) through NSERC (FONCER) initiative.

# Appendix A. Supplementary Nomenclature

The following nomenclature is useful for the appendices, in addition to the nomenclature presented at the beginning of the paper.

**Supplementary nomenclature**

| | |
|---|---|
| $q_{e,m}$ | Charge $[C]_e$, $[A\,m]_m$ |
| $d_{e,m}$ | Dipole charge separation [m] |
| $r$ | Distance from an electric charge [m] |
| $\varepsilon_0$ | Permittivity of free space [F m$^{-1}$] |
| $\mu_0$ | Permeability of free space [N A$^{-2}$] |
| $\rho_{e,m}$ | Density of charge $[C\,m^{-3}]_e$, $[A\,m^{-2}]_m$ |
| $J$ | Electric current [C s$^{-1}$] |
| $\nabla \cdot$ | Divergence operator |
| $\nabla \times$ | Curl operator |

# Appendix B. Monopoles and Dipoles

## B.1. Electric monopoles

Static electric monopoles are the most primitive elements that generates an electrical field, and they can be positive or negative. The positive charges generate an outgoing electric field and a positive potential, while the negative charges generate an ingoing electric field and a negative potential. This is shown on Fig. 17, where the color scale is the normalized value of the electric potential $\mathcal{V}_e$ and the arrows represent the electric field $\mathcal{E}_e$. In our 3D universe, the values of the potentials and fields of static charges are given by equation (14) [18–20]. However, the current paper will not limit itself to the 3D equations of electromagnetism, and more general equations will be developed.

$$\mathcal{V}_e = \frac{q_e}{4\pi\varepsilon_0 \|r\|}$$

$$\mathcal{E}_e = \frac{q_e}{4\pi\varepsilon_0 \|r\|^2} \hat{r} \qquad (14)$$

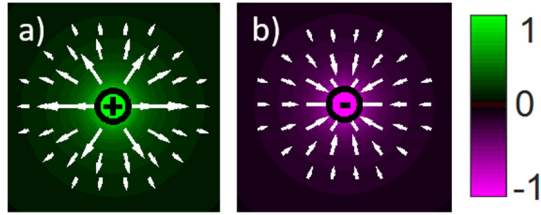

**Fig. 17.** Static electric potential and field of a: (a) positive monopole. (b) negative monopole

The color-bar used for the potential is shown on Fig. 1, but will be omitted in many other images for concision. It is normalized so that the value "1" is associated to the maximum potential and "−1" is associated to the maximum negative potential.

When we deal with more than one particle, then the total potential and field is the sum of all the individual potentials and fields, as given by equation (16) [18–20]. It should be noted that the total potential is a simple scalar sum, while the total field is a vector sum.

$$\mathcal{V}_e^{tot} = \sum_i^n \mathcal{V}_e^i, \qquad \mathcal{E}_e^{tot} = \sum_i^n \mathcal{E}_e^i \tag{15}$$

*B.2. Electric dipoles*

An electric dipole is created by placing a positive charge near a negative charge. This generates an electric potential that is positive on one side (positive pole), negative on the other side (negative pole) and null in the middle. The charge separation $d_e$ a vector corresponding to the displacement from the positive charge to the negative charge, and is mathematically defined at equation (16) [20].

$$\boldsymbol{d}_e = \boldsymbol{r}_{e+} - \boldsymbol{r}_{e-} \tag{16}$$

The electric field will then have a preferential direction along the vector $\boldsymbol{d}_e$ by moving away from the positive charge, but it will loop back on the sides to reach the negative charge. Many examples of electric dipoles are presented at Fig. 18, with the simplest form being composed of 2 opposite charges. On this figure, we notice that staking multiple dipoles in a chain will not result in a stronger dipole, because all the positive and negative charges in the middle will cancel each other. Therefore, stacking the dipoles in series will only place the poles further away from each other. However, stacking the dipoles in parallel will result in a stronger potential and field on each side of the dipole. It is also possible to see that the field will be almost perpendicular to the line of parallel dipoles, but it is an outgoing field on one side and an ingoing field on the other.

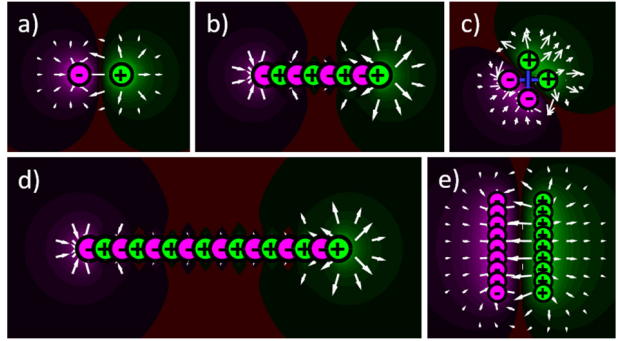

**Fig. 18.** Electric Potential and field for static monopoles placed as: (a) A simple dipole. (b) A small chain of simple dipoles. (c) A horizontal and a vertical dipole, equivalent as 2 dipoles at 45°. (d) A long chain of simple dipoles in series. (e) A long chain of simple dipoles in parallel.

To calculate the total electric potential and field of any kind of dipole, it is possible to use the equation (14), without forgetting to change the sign of $q_e$ accordingly. This sign change leads to a potential that diminishes a lot faster for dipoles at Fig. 18 when compared to the monopoles at Fig. 17. In a 3D world, with $\theta = 0$ alongside vector $\boldsymbol{d}_e$, the dipole potential will vary according to $\mathcal{V}_{dip} \propto \cos(\theta)/\|\boldsymbol{r}\|^2$, compared to the monopole potential which varies in proportion to $\mathcal{V}_e \propto 1/\|\boldsymbol{r}\|$ [19, 20].

Another important aspect of dipoles is that when $\boldsymbol{d}_e$ is small, the potential of a diagonal dipole is calculated by the linear combination of a horizontal and a vertical dipole. The potential of a dipole at angle $\theta$ ($\mathcal{V}_{dip}^\theta$) is approximated by equation (17) [19, 20]. This is easy to prove by using the previous statement that $\mathcal{V}_{dip} \propto \cos(\theta)$.

$$\mathcal{V}_{dip}^{\theta} \approx \mathcal{V}_{dip}^{x} \cos(\theta) + \mathcal{V}_{dip}^{y} \sin(\theta) \tag{17}$$

*B.3. Magnetic charges and dipoles*

Electricity and magnetism are 2 concepts with an almost perfect symmetry between them, and will lead to similar mathematical equations. First, a magnetic dipole is what is commonly called a "magnet", and is composed of a north pole (N) and a south pole (S). When compared to the electrical dipole, the north pole is mathematically identical to the positive pole and the south pole is identical to the negative pole. Therefore, the potentials and fields of magnetic dipoles are identical to those of Fig. 2, and the equations are the same as those defined by equations (14), except for the constants.

One can also mathematically define a magnetic monopole the same way as the electric monopole was defined. Although magnetic monopoles are not found in nature, nothing prevents us from using their mathematical concepts for computer vision.

**Appendix C. Mathematical Laws of EM**

*C.1. Maxwell's Equations*

The development of traditional electromagnetism was completed by J.C. Maxwell and allows to explain all the EM phenomenon using 4 mathematical equations, known as Maxwell's equations (MEq), which can be written with integrals or differential form. The first MEq is the Gauss law presented at equation (18) [18, 20]. It means that the electric field that leaves a certain volume is directly proportional to the total charge inside it. The second MEq is Gauss law of magnetism and is presented at equation (19) [18, 20]. It is identical to equation (18), except that the charge density is zero due to the inexistence of magnetic charges.

$$\nabla \cdot \boldsymbol{\mathcal{E}}_e = \frac{\rho_e}{\varepsilon_0} \tag{18}$$

$$\nabla \cdot \boldsymbol{\mathcal{E}}_m = 0 \tag{19}$$

The following MEq are known as Faraday's law (20) and Ampere's law (21). They allow to understand the behavior of EM when there are time variations of the fields [18, 20].

$$\nabla \times \boldsymbol{\mathcal{E}}_e = -\frac{\partial \boldsymbol{\mathcal{E}}_m}{\partial t} \tag{20}$$

$$\nabla \times \boldsymbol{\mathcal{E}}_m = \mu_0 \left( J + \varepsilon_0 \frac{\partial \boldsymbol{\mathcal{E}}_e}{\partial t} \right) \tag{21}$$

Another important concept in EM is the electrostatic potential $\mathcal{V}_{e,m}$, which is a scalar defined as the line integral of the field $\boldsymbol{\mathcal{E}}_{e,m}$, given by equation (22) [20].

$$\mathcal{V}_{e,m} = -\int_C \boldsymbol{\mathcal{E}}_{e,m} \cdot \mathrm{d}\boldsymbol{l} \tag{22}$$

*C.2. Adaptation of Maxwell's equations for Computer Vision*

The important equations of EM were developed in the previous section, but their current form is not adapted for computer vision. First, the presence of the constants $\varepsilon_0$ and $\mu_0$ are not useful for the current application. It is also possible to ignore the fact that magnetic charges cannot exist and regroup equations (18) and (19) to generate equation (20). Furthermore, the time variation and the current from equations (20) and (21) are ignored to generate equation (24). Thus, the 4 MEq are simplified into 2 new equations for static electromagnetism given by (23) and (24). These equations are the same for Electricity and Magnetism. Also, there is no interaction between a static electric field and a static magnetic field [18, 19]. For these reasons, the current paper will often use the term "electric" when using monopoles and "magnetic" or "magnetize" when using dipoles, because it is more intuitive.

The equation (23) means that the total virtual field going out of a surface is directly proportional to the number of virtual charges contained inside. For dipoles, the total virtual field is null because the sum of charges is always zero. The equation (24) means that there are no curl to the field. By using equation (24) with equation (22), it is possible to demonstrate equation (25) [20], which states that the field is given by the gradient of the potential.

$$\nabla \cdot \boldsymbol{E}_{e,m} = \rho_{e,m} \tag{23}$$

$$\nabla \times \boldsymbol{E}_{e,m} = 0 \tag{24}$$

$$\boldsymbol{E}_{e,m} = -\nabla V_{e,m} \tag{25}$$

With these equations demonstrated, the next step is to determine the potential and the fields that are generated by charged particles. By using equation (23) and by knowing that, at a certain radius, the field around a single particle is uniformly spread, it is possible to show that equation (3) holds. The variable "$n$" denotes the dimension of the universe where the potentials and fields are used. This means that for a 3D universe we have $\boldsymbol{E}_{e,m} \propto 1/|r|^2$, for a 2D universe we have $\boldsymbol{E}_{e,m} \propto 1/|r|^1$, while for a 1D universe, $\boldsymbol{E}_{e,m}$ is constant. This is in concordance with the real laws of electromagnetism, if we acknowledge that a 2D universe is when the infinite wire approximation is used, and that a 1D universe is when an infinite plate approximation is used. However, the current paper will not limit equation (3) to a finite universe by choosing non-integer values for $n$. Therefore, we have that $\boldsymbol{E}_{e,m} \propto 1/|r|^{n-1}$, for a universe of $n$ spatial dimensions.

**Appendix D. Geometrical Interpretation of Maxwell's Equations**

*D.1. Closed shapes with particles on the contour*

If we have a closed circle composed of charged particles on the contour, then equation (23) allow to see that the field will be almost null inside the circle, but really high as soon as we are outside the circle. The potential is strong and constant in the middle of the circle, but diminishes rapidly outside the circle [19, 20], as observed at Fig. 19. This is because the potential is scalar, therefore the contribution of each particle will be summed. However, the field requires a vector sum, which means that they will cancel each other in the middle as the vectors will be of opposite direction, but they will add themselves outside of the circle. This holds true for any kind of closed shape, although the perfect symmetry of a circle makes the cancelation of the field more effective.

By using circle of dipoles instead of monopoles, with the radii being $R_-$ and $R_+ > R_-$, then the field will be null and the potential almost constant everywhere, except for a position $R$ that respects the inequality $R_- < R < R_+$. This is due to the gauss law (23) and is observed at Fig. 19. It also holds true for any kind of closed shape with dipoles on all of its contour.

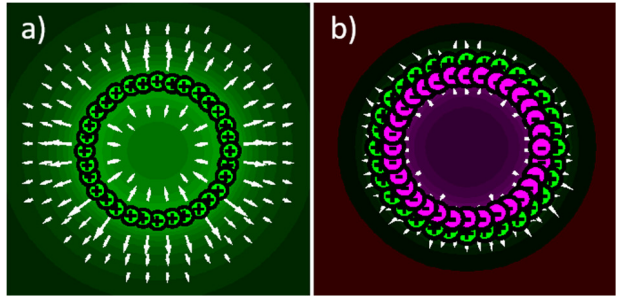

**Fig. 19.** Potential and field for circles, with $n = 3$, for (a) positive monopoles. (b) parallel dipoles.

*D.2. Corners with particles on the contour*

If we have a corner composed of charged particles as presented at Fig. 20, then it is possible to analyse the potential and field and to see that it somewhat resembles the closed circle. The concave part of the corner will have a low field due to the vector sum of opposite vectors, but it will be the point with the highest potential, just like the inside of the circle. The convex part of the corner will have a slightly higher field because the vectors are less destructive, but the potential will be a lot lower because it is further away from the other charges, similarly to the exterior of the circle. Finally, the flat parts of the corner will have the highest field, but average potential.

At Fig. 20, it is possible to see that the field on the corners will have a diagonal direction due to the contribution of both sides of the corner. For the dipole corner, the behavior is really similar than the monopole corner, except that the concave part has an ingoing field with negative potential, while the convex part has and outgoing field with positive potential. It should be noted that the field will tend to be at an angle of 45° when it is far at the top-left or bottom-right of the corner.

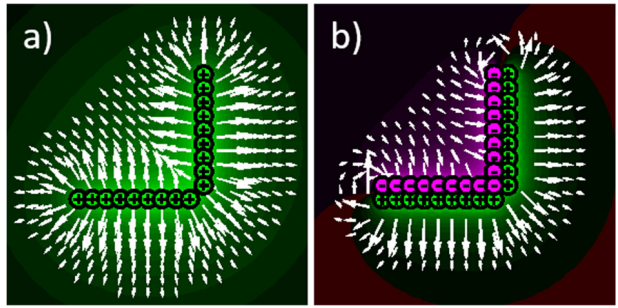

**Fig. 20.** Potential and field for corners of (a) positive monopoles. (b) parallel dipoles.

## Appendix E. Stroke Manipulations

This section presents algorithms on how to manipulate the strokes or contours of an image and how to grow/shorten specific regions from the contour. The pseudocodes make use of some Matlab® functions for computer vision, but they all have their equivalent in other image processing libraries such as OpenCV®.

*E.1. Grow and unite contour regions*

A contour region is defined as a group of pixel that are part of the contour. For example, the high potential region will be everywhere on the contour with a high value of $V_e$. However, due to discretization, the regions might have discontinuities. Also, it could be required to simply grow the desired regions by a specific number of pixels at each side of the region, but by keeping it on the contour.

24In the current paper, the regions are always grown by a percentage of the biggest dimension of an image (%BL). This allows to always be consistent no matter the resolution or the scale of the image. The regions are expended using a loop of image dilation. The dilations will expand the region one pixel in all directions, and then be multiplied by the contours to remove the undesired growth. This process is detailed in Algorithm 2.

Algorithm 1. Pseudocode for growth of a region on the contour

```
// growthPercentage: Desired %BL for the growth
// regionOnContour: Matrix with value 1 on parts of the contour to grow
// contour: Matrix with value 1 on the contour
Function GROW_REGION(growthPercentage, regionOnCoutour, contour)
// Find the number of pixels to grow
numberPixelToGrow = round(growthPercentage * max(size(image)));
// Compute the geodesic distance between each point on the contour and the region
// "bwdistgeodesic" is a MATLAB function that computes the geodesic distance
geodesicDistance = bwdistgeodesic(contour, regionOnContour);

// Define the grown region as the region with distance lower then the threshold
grownRegion = (geodesicDistance <= numberPixelToGrow);
RETURN grownRegion;
```

An important application for this region growth is to be able to unite pixels that are near each other into a single region. Due to the discretization, some regions of high potential will be broken into multiple but nearby pixels. By using the growth technique that was presented, the pixels will unite to form a solid region on the contour.

*E.2. Finding Stroke Orientation*

Finding the orientation of a stroke is crucial to the use of directional magnetic convolutions. The way to do it, is to start from the extremity of the stroke, and loop every pixel from this point to find the angle for the next point, as depicted in Algorithm 3. Since the image is in a matrix, each pixel has 8 possible neighbors, meaning the angles will always be $\theta = n \cdot \frac{\pi}{4}$, with the value of $n = [1,2,...,8]$. However, the Algorithm 3 applies multiple consecutive smoothing on the delta values, meaning the angle will have a lot more than 8 possible values.

Algorithm 2. Pseudocode for finding the orientation of a stroke

```
// stroke: Matrix with value 1 on a stroke, and 0 elsewhere
Function STROKE_ORIENTATION(stroke)
// Make sure the region is thin, with only 2 neighbours everywhere, except on
intersections
strokeThin = MorphologicalThinning(stroke);

// Remove the intersections from the stroke, which creates more strokes
kernel = [1,1,1; 1,0,1; 1,1,1];
convol2D = convolution2D(strokeThin, kernel);
hasLessThan3Neighbours = convol2D < 3;
strokeThinNoIntersect = strokeThin ∘ hasLessThan3Neighbours;

FOR EACH subStroke IN strokeThinNoIntersect
    // If the substroke is open, start from the Extremety
    // Else, choose a random point as the Extremety, and remove one of its neighbour
```

```
      isOpen = any(convol2D < 2)
      IF subStroke IS isOpen
         isExtremety = convol2D == 1;
         Extremety = chooseRandomPoint(isExtremety);
      ELSE
         Extremety = chooseRandomPoint(subStroke);
         subStroke = removeOnePointNearAfromB(Extremety, subStroke);
      ENDIF

      // Initialize loop parameters
      numPixelsRemaining = count(subStroke > 0);
      subStrokeRemaining = subStroke;
      currentPoint = Extremety;
      allDeltaX = MatrixOfNAN;
      allDeltaY = MatrixOfNAN;

      // Loop all the points of the stroke from the Extremety
      WHILE numPixelsRemaining > 0
         nextPoint = findNearestPointFromPointOnStroke(currentPoint,
subStrokeRemaining);
         allDeltaX (currentPoint) = XDistanceBetween(currentPoint, nextPoint);
         allDeltaY (currentPoint) = YDistanceBetween(currentPoint, nextPoint);
         numPixelsRemaining = count(subStroke > 0);
      ENDWHILE
ENDFOR

allAngles = ATAN2(allDeltaY, allDeltaX);

RETURN allAngles;
```

2610. Li, B., Acton, S.T.: Vector Field Convolution for Image Segmentation using Snakes. In: 2006 International Conference on Image Processing. pp. 1637–1640 (2006).
11. Sun, G., Liu, Q., Liu, Q., Ji, C., Li, X.: A novel approach for edge detection based on the theory of universal gravity. Pattern Recognit. 40, 2766–2775 (2007).
12. Lenz, I., Lee, H., Saxena, A.: Deep learning for detecting robotic grasps. Int. J. Robot. Res. 34, 705–724 (2015).
13. Russell, S.J., Norvig, P.: Artificial Intelligence: A Modern Approach. Prentice Hall (2010).
14. Gurney, K.: An Introduction to Neural Networks. CRC Press (2003).
15. MNIST handwritten digit database, Yann LeCun, Corinna Cortes and Chris Burges, http://yann.lecun.com/exdb/mnist/.
16. Salman, N.: Image Segmentation Based on Watershed and Edge Detection Techniques. (2004).
17. Davis, R.S.: Optical images by quadrupole convolution, http://www.google.com.uy/patents/US5027419, (1991).
18. Maxwell, J.C.: A Treatise on Electricity and Magnetism. Clarendon Press (1881).
19. Feynman, R.P., B, F.R.P.S.M.L.L.R., Leighton, R.B., Sands, M.: The Feynman Lectures on Physics, Desktop Edition Volume II: The New Millennium Edition. Basic Books (2013).
20. Rothwell, E.J., Cloud, M.J.: Electromagnetics, Second Edition. CRC Press (2008).
21. Costa, L. da F.D., Cesar, R.M., Jr.: Shape Analysis and Classification: Theory and Practice. CRC Press, Inc., Boca Raton, FL, USA (2000).
22. Loncaric, S.: A survey of shape analysis techniques. Pattern Recognit. 31, 983–1001 (1998).
23. Calli, B., Wisse, M., Jonker, P.: Grasping of unknown objects via curvature maximization using active vision. In: 2011 IEEE/RSJ International Conference on Intelligent Robots and Systems (IROS). pp. 995–1001 (2011).
24. Kuhl, F.P., Giardina, C.R.: Elliptic Fourier features of a closed contour. Comput. Graph. Image Process. 18, 236–258 (1982).
25. Backes, A.R., Bruno, O.M.: A Graph-Based Approach for Shape Skeleton Analysis. In: Image Analysis and Processing – ICIAP 2009. pp. 731–738. Springer, Berlin, Heidelberg (2009).
26. Pratt, I.: Shape Representation Using Fourier Coefficients of the Sinusoidal Transform. J. Math. Imaging Vis. 10, 221–235 (1999).
27. Smach, F., Lemaître, C., Gauthier, J.-P., Miteran, J., Atri, M.: Generalized Fourier Descriptors with Applications to Objects Recognition in SVM Context. J. Math. Imaging Vis. 30, 43–71 (2008).
28. Corke, P.: Robotics, Vision and Control: Fundamental Algorithms in MATLAB. Springer Science & Business Media (2011).
29. Fornberg, B.: Generation of finite difference formulas on arbitrarily spaced grids. Math. Comput. 51, 699–706 (1988).
30. Malik, J., Perona, P.: A computational model of texture segmentation. In: , IEEE Computer Society Conference on Computer Vision and Pattern Recognition, 1989. Proceedings CVPR '89. pp. 326–332 (1989).